%% file: root.tex
\documentclass[letterpaper, 10 pt, conference]{ieeeconf}  

\IEEEoverridecommandlockouts                              

\title{\LARGE \bf
Learning Symbolic Operators for Task and Motion Planning
}

\author{\textbf{Tom Silver$^*$, Rohan Chitnis$^*$, Joshua Tenenbaum, Leslie Pack Kaelbling, Tom\'{a}s Lozano-P\'{e}rez}\\\textnormal{MIT Computer Science and Artificial Intelligence Laboratory}\\\texttt{\{tslvr, ronuchit, jbt, lpk, tlp\}@mit.edu}\thanks{\hspace*{-1em} $^*$ Equal contribution.}}

\input{preamble.tex}

\input{variables.tex}

\begin{document}

\bibliographystyle{IEEEtran}

\maketitle
\thispagestyle{empty}
\pagestyle{empty}

\begin{abstract}
\input{abstract.tex}
\end{abstract}

\input{intro.tex}
\input{related.tex}
\input{setting.tex}

\input{approach1.tex}
\input{approach2.tex}
\input{experiments.tex}
\input{conclusion.tex}
\input{acknowledgements.tex}

\addtolength{\textheight}{-9cm}   





\bibliography{references}

\end{document}

%% file: preamble.tex
\usepackage{multicol}
\usepackage{color}
\usepackage{amssymb}
\usepackage{amsmath}
\usepackage{tikz}
\usepackage{chngcntr}
\usepackage[makeroom]{cancel}
\usetikzlibrary{arrows,decorations.markings}

\usepackage{hyperref}

\usepackage{booktabs}
\usepackage{graphicx}
\usepackage{mathtools}
\usepackage{bbm}
\usepackage[switch]{lineno}

\usepackage[font=small]{caption}

\usepackage{algorithm2e}
\let\oldnl\nl
\newcommand{\nonl}{\renewcommand{\nl}{\let\nl\oldnl}}

\usepackage{indentfirst}

{\begin{list}{$\bullet$}{%
    \setlength{\topsep}{0in}
    \setlength{\partopsep}{0in}
    \setlength{\itemsep}{0in}
    \setlength{\parsep}{0in}
    \setlength{\leftmargin}{1.5em}
    \setlength{\rightmargin}{0in}
}
}%
{\end{list}
}

\newcommand{\secref}[1]{Section~\ref{#1}}

\newcommand{\figref}[1]{Figure~\ref{#1}}
\newcommand{\algref}[1]{Algorithm~\ref{#1}}

\newcommand{\tabref}[1]{Table~\ref{#1}}

\newtheorem{defn}{Definition}

\counterwithin*{thm}{section}

\def\thickhline{%
  \noalign{\ifnum0=`}\fi\hrule \@height \thickarrayrulewidth \futurelet
   \reserved@a\@xthickhline}
\def\@xthickhline{\ifx\reserved@a\thickhline
               \vskip\doublerulesep
               \vskip-\thickarrayrulewidth
             \fi
      \ifnum0=`{\fi}}

\newlength{\thickarrayrulewidth}

%% file: variables.tex
\newcommand{\X}{\mathcal{X}}

\newcommand{\tamp}{{\sc tamp}}
\newcommand{\loft}{{\sc loft}}

\newcommand{\D}{\mathcal{D}}

\DeclarePairedDelimiterX{\infdivx}[2]{(}{)}{%
  #1\;\delimsize\|\;#2%
}

%% file: abstract.tex
Robotic planning problems in hybrid state and action spaces can be solved by integrated task and motion planners (\tamp{}) that handle the complex interaction between motion-level decisions and task-level plan feasibility. 
\tamp{} approaches rely on domain-specific symbolic operators to guide the task-level search, making planning efficient.
In this work, we formalize and study the problem of operator learning for \tamp{}.
Central to this study is the view that operators define a lossy abstraction of the transition model of a domain.
We then propose a bottom-up relational learning method for operator learning and show how the learned operators can be used for planning in a \tamp{} system.
Experimentally, we provide results in three domains, including long-horizon robotic planning tasks. We find our approach to substantially outperform several baselines, including three graph neural network-based model-free approaches from the recent literature.
Video: \url{https://youtu.be/iVfpX9BpBRo}. Code: \url{https://git.io/JCT0g}

%% file: intro.tex
\section{Introduction}
\label{sec:intro}

\begin{figure}[t]
  \centering
  \noindent
    \includegraphics[width=0.95\columnwidth]{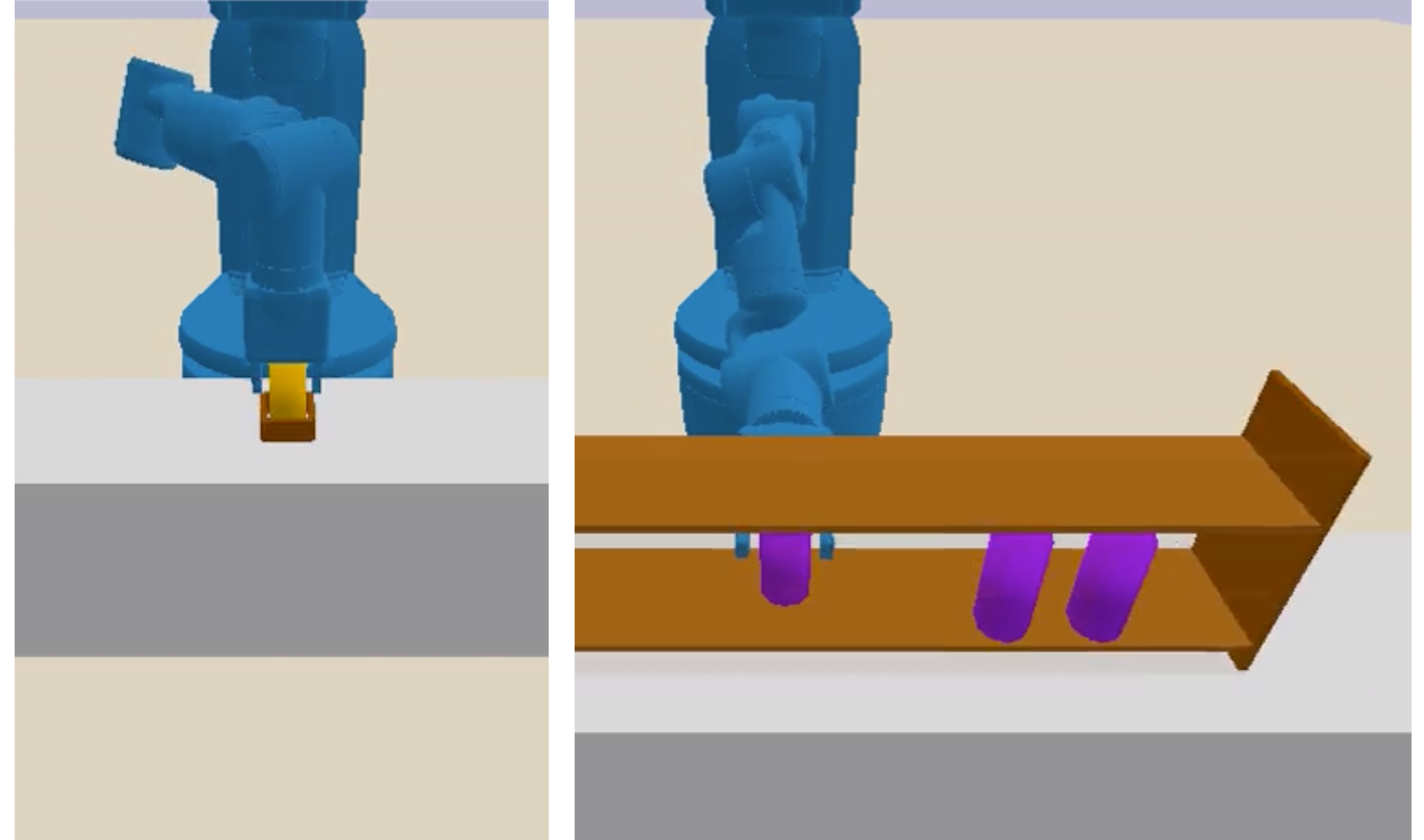}
    \caption{Snapshots from the ``Painting'' domain (\secref{sec:experiments}). \emph{Left:} An object that is grasped from the top can be placed into a small open-faced box. \emph{Right:} A side grasp is required to place an object into a shelf with a ceiling. The chosen continuous grasp argument for the \texttt{Pick} action therefore later influences the effects of placing.}
  \label{fig:teaser}
\end{figure}

Robotic planning problems are often formalized as \emph{hybrid} optimization problems, requiring the agent to reason about both discrete and continuous choices (e.g., \emph{Which object should I grasp?} and \emph{How should I grasp it?})~\cite{tampsurvey}.
A central difficulty is the complex interaction between low-level geometric choices and high-level plan feasibility.
For example, how an object is grasped affects whether or not it can later be placed into a shelf (\figref{fig:teaser}).
Task and motion planning (\tamp{}) combines insights from AI planning and motion planning to address these challenges~\cite{tampsurvey,gravot2005asymov,tampconstraints,tamphpn,tampinterface,tamplgp}.
\tamp{} uses symbolic planning operators to search over symbolic plans, biasing the search over motions.
Operators are hand-specified in all popular \tamp{} systems~\cite{zhu2020hierarchical}, requiring expert input for each domain. Instead, we aim to develop a domain-independent operator learning algorithm for \tamp{}.

Symbolic operators are useful for \tamp{} in three key ways.
First, operators let us efficiently determine that many plans have zero probability of reaching a goal, regardless of the choice of continuous action parameters, allowing us to ignore such plans in the search.
Second, operators permit a bilevel optimization approach, with symbolic planning providing a dense sequence of subgoals for continuous optimization~\cite{tampsurvey,tampinterface}.
Third, explicit PDDL-style operators~\cite{pddl} allow us to automatically derive AI planning heuristics, which dramatically speed up the search over symbolic plans.

In this work, we formalize and study the problem of learning operators for \tamp{} from data, using a bottom-up relational learning method.
We consider a setting with a deterministic low-level environment simulator, a set of hybrid controllers with associated samplers for the continuous parameters, an object-oriented continuous state, and a set of predicates that collectively define a lossy abstraction of the low-level state.
In this setting, planning is possible \emph{without} operators via breadth-first search over sequences of controllers and a schedule for calling the associated samplers (\secref{sec:experiments}, Baseline 5).
However, we find that making use of the operators within \tamp{} provides enormous benefits.

Our approach continues a line of recent work that seeks to exploit properties of task distributions to make \tamp{} more efficient.
In particular, our approach can be understood as a model-based method for learning guidance in hybrid planning problems: the operators define an abstract transition model that provides guidance.
Recent literature on learning for \tamp{} considers various model-free counterparts. Kim and Shimanuki~\cite{kim2020learning} learn an ``abstract Q-function'' that implicitly defines a goal-conditioned policy over symbolic actions, while Driess et al.~\cite{driess2020deep} learn a recurrent model that directly produces symbolic plans.
In our experiments, we consider three model-free baselines inspired by these methods.

In this paper, we propose the Learning Operators For \tamp{} (\loft{}) algorithm, and make the following contributions: (1) we formalize the problem of operator learning in \tamp{}; (2) we propose a relational operator learning method, and show how to use the learned operators to quickly solve \tamp{} problems; (3) we provide experimental results in three domains, including long-horizon robotic planning tasks, that show the strength of our approach over several baselines.

%% file: related.tex
\section{Related Work}
\label{sec:related}

\subsection{Task and Motion Planning (\tamp{})}

The field of \tamp{} emerged from the combination of AI methods for task planning~\cite{bonet2001planning} and robotic methods for motion planning. \tamp{} methods are focused on solving multimodal, continuous robotic planning problems in highly unstructured environments~\cite{gizzi2019creative,sarathy2020spotter}; see Garrett et al.~\cite{tampsurvey} for a recent survey. Approaches to \tamp{} can be broadly categorized based on how the treatment of symbolic reasoning interacts with the treatment of continuous variables during planning. Some approaches to \tamp{} involve optimization over trajectories~\cite{tampconstraints,tamplgp}, while others use sampling-based procedures~\cite{tamphpn,tampinterface,pddlstream}. In all cases, popular \tamp{} systems rely on hand-specified planning models (e.g., PDDL operators), a limitation we aim to address in this paper.

\subsection{Learning for Task and Motion Planning}

Learning techniques have been integrated into many aspects of \tamp{} systems, from learning samplers for continuous values~\cite{wang2018active,ltampchitnis,ltampkim1,chitnis2019learning} to learning guidance for symbolic planning~\cite{kim2020learning,driess2020deep,ltampchitnis}. The latter is our focus in this paper; we assume samplers are given, and we aim to learn operators that enable symbolic planning in \tamp{}.

Most relevant to our work are efforts to learn model-free search guidance for symbolic planning \cite{kim2020learning,driess2020deep}.
A challenge in applying model-free techniques in the \tamp{} setting is that there is no obvious way to ``execute'' an action in the space of symbolic transitions.
Kim and Shimanuki~\cite{kim2020learning} address this challenge by sampling low-level transitions at each step during symbolic planning; Driess et al.~\cite{driess2020deep} instead learn a recurrent ``Q-function'' that takes in a sequence of actions and an \emph{initial} state.
In our experiments, we consider three baselines inspired by these model-free approaches.

\subsection{Learning Symbolic Operators}

Learning symbolic planning operators
has a very long history in the planning literature; see Arora et al.~\cite{arora2018review} for a comprehensive review. While it has been studied for decades~\cite{zpk,rodrigues2011active,cresswell2013acquiring,aineto2018learning,xia2018learning}, operator learning has not been studied in the \tamp{} setting, where the learned operators must be understood as a lossy and abstract description of a low-level, geometric planning problem. In \tamp{}, the operators are a means to an end, not the entire story: the operators enable symbolic planning, which in turn produces candidate symbolic plans for a low-level optimizer. The operators in \tamp{}, therefore, are useful in that they give guidance to the overall planning procedure.

%% file: setting.tex
\section{Problem Setting}
\label{sec:setting}

We consider a standard \tamp{} setting with low-level states $x \in \X$, where $x$ is a mapping from a set of typed objects $O$ to attributes and their corresponding values. The attributes are fixed for each object type, and all values are real-valued vectors.
For example, $x$ may include the continuous 6D pose of each object in the scene.
We are given a set of hybrid controllers $\Pi = \{\pi_1, \dots, \pi_k\}$, each parameterized by zero or more discrete objects $\overline{o} = (o_1, \dots, o_m)$ with $o_i \in O$, and by a real-valued continuous vector parameter $\overline{\theta}$. An action $a$ is an instantiation of a controller with arguments, both discrete and continuous. For example, the action \texttt{Pick(obj1$, \theta$)} is a call to a controller $\texttt{Pick} \in \Pi$ that will attempt to pick the object $o_1 \gets$ \texttt{obj1} using grasp arguments $\overline{\theta}\gets\theta$.

Each controller is associated with a given \emph{sampler} for the continuous parameters $\overline{\theta}$, conditioned on the low-level state and discrete arguments. Samplers produce values that satisfy implicit constraints specific to their controller; for example, the sampler for the \texttt{Pick} controller produces feasible grasps $\theta$ based on the state $x$ and the discrete argument $o_1$.\footnote{Not all \tamp{} systems use samplers; some are optimization-based~\cite{tamplgp}. Our approach is not limited to sampling-based \tamp{} systems, but we find it convenient for exposition to describe our problem setting this way.}

We are given a low-level simulator $f$ defining the environment dynamics. $f$ maps a low-level state and an action to a next low-level state, denoted $x_{t+1} = f(x_t, a_t)$. We do not assume any analytical knowledge of the transition model.

We assume access to a set of predicates $P$. Each predicate $p(\overline{o}) \in P$ represents a named relation among one or more objects in $O$.
For example, \texttt{On($y, z$)} encodes whether an object $y$ is on top of another object $z$.
In this work, all predicates are discrete: arguments are objects, and predicates either hold or do not hold (\emph{cf.} numeric fluents).
A predicate with variables as arguments is \emph{lifted}; a predicate with objects as arguments is \emph{ground}. Each predicate is a classifier over the low-level state $x$. Given a state $x$, we can compute the set of all ground predicates that hold in the state, denoted $s = \textsc{Parse}(x)$, where $\textsc{Parse}$ is a deterministic function.
For example, $\textsc{Parse}$ may use the geometric information in the low-level state $x$ to determine which objects are on other objects, adding \texttt{On($y, z$)} to $s$ if object $y$ is on object $z$.
We refer to $s$ as the \emph{symbolic state}.
The assumption that predicates are provided is limiting but standard in the learning-for-\tamp{} literature~\cite{kim2020learning,driess2020deep,wang2018active}.
We emphasize that the predicates are \emph{lossy}, in the sense that transitions at the symbolic level can be non-deterministic, even though the low-level simulator $f$ and \textsc{Parse} function are deterministic~\cite{marthi2007angelic}.

We are given a set of planning problems $\{(O, x_0, G)\}$ to solve. Here, $O$ is an object set, $x_0 \in \X$ is an initial low-level state, and $G$ is a goal. All problems share a simulator $f$, predicates $P$, and controllers $\Pi$.
A goal $G$ is a (conjunctive) set of ground predicates over the object set $O$; we say $G$ is \emph{achieved} in state $x$ if $G \subseteq \textsc{Parse}(x)$.
A solution to a planning problem is a \emph{plan}: a sequence of actions $(a_0, \ldots, a_{T-1})$ where $x_{t+1} = f(x_t, a_t)$ and $x_T$ achieves $G$.

Since we are interested in learning-based approaches, we suppose that we are given a dataset $\D$, collected offline, of low-level transitions $(x_i, a_i, x_{i+1}, G_i)$ generated from planning in (typically smaller) problems from the same family. See \secref{sec:experiments} for details on how we collect $\D$ in practice.

%% file: approach1.tex
\section{Task and Motion Planning with Operators}
\label{sec:approach1}

Most \tamp{} systems rely on hand-defined, domain-specific symbolic \emph{operators} to guide planning.
In this section, we define operators and describe how they are used for planning in ``search-then-sample'' \tamp{} methods~\cite{tampsurvey}.

A symbolic operator is composed of a controller, parameters, a precondition set, and an effect set.\footnote{Each controller can be associated with \emph{multiple} operators.
For example, in the Painting domain shown in \figref{fig:teaser}, a generic \texttt{Place} controller would have two operators, one for placing into the shelf (with \texttt{HoldingSide} in the preconditions and \texttt{InShelf} in the effects) and another for placing into the box (with \texttt{HoldingTop} and \texttt{InBox} respectively).}
The parameters are typed placeholders for objects that are involved in the discrete controller parameters, the precondition set, or the effect set.
Preconditions are lifted predicates over the parameters that describe what must hold for the operator to be applicable.
Effects are (possibly negated) lifted predicates over the parameters that describe how the symbolic state changes as a result of applying this operator (executing this controller).
The operator can be \emph{grounded} by assigning the parameters to objects, making substitutions in the parameters, preconditions, and effects accordingly.
In this paper, operators do not model the influence of the continuous action parameters, and therefore make predictions based on the discrete controller parameters and symbolic state alone.
See the third panel of Figure \ref{fig:pipeline} for an example operator (\texttt{Pick0}).

To understand how operators can be used to guide \tamp{}, we turn to the following definitions.

\begin{defn}[Action template]
\label{defn:actiontemplate}
An \emph{action template} is a controller $\pi(\overline{o}, \overline{\theta})$ and an assignment of the controller's discrete parameters $\overline{o} \gets o$, with the continuous parameters left unassigned. We denote the action template as $\pi(o, \cdot)$.
\end{defn}

An action template can be understood as an action with a ``hole'' for the continuous arguments of the controller.
For example, \texttt{Pick(obj1$, \cdot$)} is an action template with a hole left for the continuous grasp arguments.

\begin{defn}[Plan skeleton]
\label{defn:planskeleton}
A \emph{plan skeleton} is a sequence of action templates $(\pi_1(o_1, \cdot), \dots, \pi_\ell(o_\ell, \cdot))$.
\end{defn}

A plan skeleton can be \emph{refined} into a plan by assigning values to all of the continuous parameters in the controller: $(\pi_1(o_1, \theta_1), \dots, \pi_\ell(o_\ell, \theta_\ell)) = (a_1, \dots, a_\ell)$.
The main role of operators is to efficiently generate plan skeletons that can be refined into a solution plan.
Given a goal $G$ and an initial low-level state $x_0$ with symbolic state $s_0 = \textsc{Parse}(x_0)$, we can use the operators to search for a plan skeleton that achieves $G$ symbolically, before needing to consider any continuous action arguments.
Importantly, though, a plan skeleton that achieves $G$ symbolically has no guarantee of being refinable into a plan that achieves $G$ in the environment.
This complication is due to the lossiness of the symbolic abstraction induced by the predicates.

To address this complication, search-then-sample \tamp{} methods perform a bilevel search, alternating between \emph{high-level symbolic planning} to search for plan skeletons and \emph{low-level optimization} to search for continuous arguments that turn a plan skeleton into a valid plan~\cite{gravot2005asymov,tampconstraints,tampinterface}. 

In this work, for high-level planning, we use $\text{A}^*$ search.
Importantly, our operators are compatible with PDDL representations~\cite{pddl}, meaning we can use classical, domain-independent planning heuristics in this search (hAdd~\cite{bonet2001planning} in all experiments).
For low-level optimization, we conduct a backtracking search over continuous parameter assignments to attempt to turn a plan skeleton into a plan.
Recall that each step in the plan skeleton is an action template, and that each controller is associated with a sampler.
For each action template, we invoke the sampler to produce continuous arguments, which in turn leads to a complete action.
The backtracking is conducted over calls to the sampler for each step of the plan skeleton.
This search terminates with success if following the action sequence in the simulator $f$ results in a low-level state that achieves the goal.
We allow a maximum of $N_\text{samples}$ calls to each step's sampler before backtracking.
If search is exhausted, control is returned to the high-level $A^*$ search to produce a new candidate plan skeleton.

A major benefit of having symbolic operators is that they allow us to prune the backtracking search by only considering plan prefixes that induce trajectories in agreement with the expected symbolic trajectory.
For example, consider a plan prefix $(a_0, a_1, a_2)$, which induces the trajectory $(x_0, x_1, x_2, x_3)$.
We can call \textsc{Parse} on each state to produce the symbolic state sequence $(s_0, s_1, s_2, s_3)$; if the transition $s_2 \to s_3$ is not possible under the operators, then the continuous arguments of $a_2$ should be resampled.

The \tamp{} algorithm described above was selected due to its relative simplicity and strong empirical performance~\cite{tampinterface}. Note, however, that it is \emph{not} probabilistically complete; it is possible that the low-level optimization cannot refine a valid skeleton within the $N_\text{samples}$ allotment, and since there is no ability to revisit a skeleton once low-level optimization fails, the algorithm may miss a solution even if one exists.

%% file: approach2.tex
\section{Learning Symbolic Operators for \tamp{}}
\label{sec:approach2}

\begin{figure*}[t]
  \centering
  \noindent
    \includegraphics[width=\textwidth]{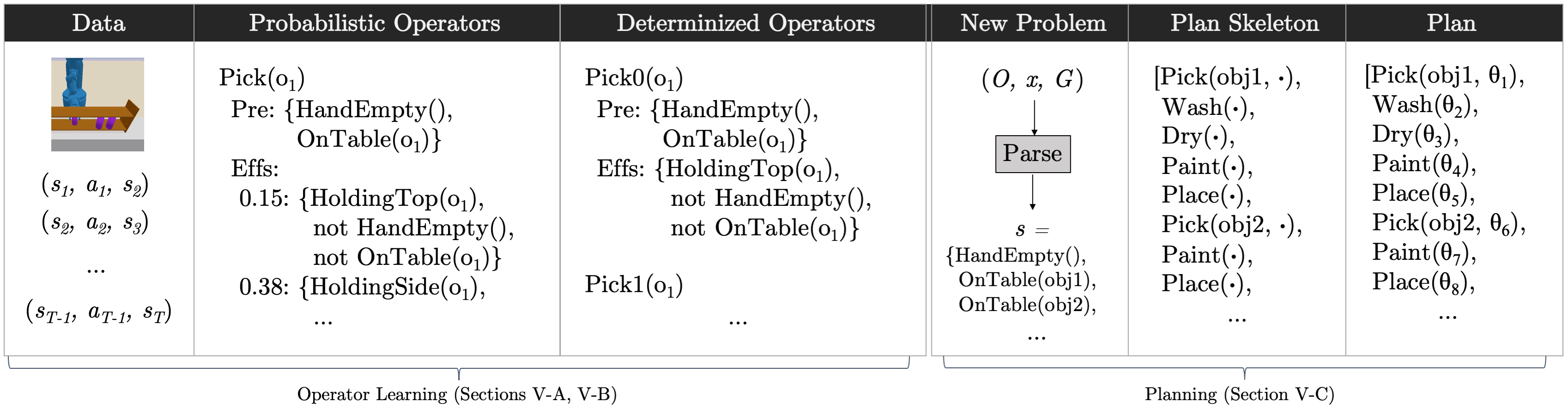}
    \caption{An example of the complete pipeline. We learn probabilistic operators on transition data, then determinize them. The operators are used to generate high-level plan skeletons for new \tamp{} problem instances, providing guidance that makes planning more efficient.}
  \label{fig:pipeline}
\end{figure*}

In this section, we describe our main approach: Learning Operators For \tamp{}, or \loft{} for short.
We begin in \secref{subsec:repr} with the observation that we can learn \emph{probabilistic} operators to account for the inevitable lossiness induced by the symbolic abstraction.
In \secref{subsec:learnndrs}, we describe an efficient algorithm for learning these probabilistic operators from transition data.
Finally, in \secref{subsec:planning} we demonstrate how to use the learned probabilistic operators for \tamp{}.
See \figref{fig:pipeline} for an example of the full pipeline.

\subsection{Probabilistic Operators as Symbolic Transition Models}
\label{subsec:repr}

From a planning perspective, operators can be seen as a substrate for guiding search.
An alternative perspective that is more amenable to learning is that the operators comprise a \emph{symbolic transition model}, describing the distribution of next symbolic states $s_{t+1}$ given a current symbolic state $s_t$ and action $a_t$: $P(s_{t+1} \mid s_t, a_t)$.\footnote{In general, the nondeterminism in the symbolic transitions may depend on the policy for generating the transitions and the low-level dynamics. For our purposes, the precise semantics are unimportant as the probabilities are only used to filter out rare outcomes, and are then discarded (\secref{subsec:planning}).}
This probabilistic framing follows from the fact that no deterministic function could accurately model these symbolic transitions, given the lossiness of the abstraction induced by $\textsc{Parse}$.

To learn operators that comprise a probabilistic transition model, we consider operators with \emph{probabilistic effects}, like those found in PPDDL~\cite{ppddl}.
These probabilistic operators are equivalent to the deterministic operators described in \secref{sec:approach1}, except that they contain a categorical distribution over effect sets rather than a single effect set. See the second panel in \figref{fig:pipeline} for an example (\texttt{Pick}).

\subsection{Learning Probabilistic Operators}
\label{subsec:learnndrs}

\input{learning_pseudocode}

We now describe a bottom-up relational method for learning probabilistic operators in the \tamp{} setting.
Recall that we are given a dataset $\D = \{(x_i, a_i, x_{i+1}, G_i)\}$ of low-level transitions.
(The goals are irrelevant for learning a transition model, but they are included in $\D$ because they prove useful for baselines in our experiments.)
These data can be converted into symbolic transitions $\{(s_i, a_i, s_{i+1})\}$ by calling the \textsc{Parse} function on the low-level states.
We now have samples from the distribution $P(s_{t+1} \mid s_t, a_t)$ that we wish to learn.
The data can be further partitioned by controller.
Let $\D_{\pi}$ denote the dataset of transitions for controller $\pi \in \Pi$, e.g., $\D_{\texttt{Pick}} = \{(s_i, a_i, s_{i+1}) \in \D : a_i = \texttt{Pick}(\cdot, \cdot)\}$.
With these datasets in hand, the algorithm for learning probabilistic operators proceeds in three steps: (1) lifted effect clustering, (2) precondition learning, and (3) parameter estimation.
See \algref{alg:learning} for pseudocode.

\emph{Lifted Effect Clustering.} We begin by clustering the transitions in $\D_{\pi}$ according to lifted effects.
For each transition $(s_i, a_i, s_{i+1})$, we compute ground effects using two set differences: $s_{i+1} - s_i$ are positive effects, and $s_{i} - s_{i+1}$ are negative effects.
We then cluster pairs of transitions together if their effects can be \emph{unified}, that is, if there exists a bijective mapping between the objects in the two transitions such that the effects are equivalent up to this mapping.
This unification can be checked in time linear in the sizes of the effect sets.
Each of the resulting clusters is labelled with the lifted effect set, where the objects in the effects from any arbitrary one of the constituent transitions are replaced with placeholders.  \algref{alg:learning} uses the notation $\D_{(\pi, \text{eff})}$ to denote the dataset for controller $\pi$ and lifted effect set ``eff.''

\emph{Precondition Learning.} Next, we learn one or more sets of preconditions for each lifted effect cluster.
We perform two levels of search: an outer greedy search over sets of preconditions, and an inner best-first search over predicates to include in each set.
The outer search is initialized to an empty set and calls the inner search to generate successors one at a time, accepting all successors until terminating after a maximum number of steps or failure of the inner search.

The inner search for a single precondition set is initialized by lifting each previous state $s_i$ for all transitions $(s_i, a_i, s_{i+1})$ in the effect cluster; each object in $s_i$ is replaced with a placeholder variable, and the resulting predicate set represents a (likely over-specialized) candidate precondition.
Successors in this inner search are generated by removing each possible precondition from the current candidate set.

A transition is considered \emph{explained} by a precondition set if there exists some substitution of the precondition variables to the objects in the transition so that the effects are equivalent under the substitution to that of the current effect cluster.
A precondition set is desirable if it leads to many ``true positive'' transitions --- ones that are explained by this precondition set, but not by any previously selected ones.
A precondition set is undesirable if it leads to many ``false positive'' transitions, where the preconditions hold under some variable substitution, but the effects do not match the cluster.
We therefore assign each candidate set a weighted sum score (higher is better): $\beta\ \times\ $(\# true positives) $-$ (\# false positives), where $\beta$ is a hyperparameter ($\beta=10$ in all experiments).
The inner search terminates after a maximum number of iterations or when no improving successor can be found.
The highest-scoring candidate is returned.

Precondition learning is the computationally hard step of the overall algorithm; the outer and inner searches are approximate methods for identifying the best sets of preconditions under the score function.
The computational complexity is bounded by the number of iterations in the inner search (100 in all experiments), the amount of data, the number of controllers, and the number of predicates in the largest state.

\emph{Parameter Estimation.} After precondition learning, we have one or more sets of preconditions for each set of lifted effects, for each controller.
Looking between lifted effects, it will often be the case that the same precondition set (up to unification) appears multiple times.
In the example of \figref{fig:pipeline} for \texttt{Pick}, the precondition set $\{$\texttt{HandEmpty()}, \texttt{OnTable($o_1$)}$\}$ may be associated with two sets of effects, one that includes \texttt{HoldingSide($o_1$)} and another that includes \texttt{HoldingTop($o_1$)}.
We combine effects with matching preconditions to initialize the probabilistic operators.

All that remains is to estimate the probabilities associated with each operator's effect sets.
This parameter estimation problem reduces to standard categorical distribution learning; for each pair of lifted preconditions and effects, we count the number of transitions for which the lifted preconditions hold and the number for which the effects follow, and divide the latter by the former.
This results in a final set of probabilistic operators, as shown in the second panel of \figref{fig:pipeline}.\footnote{These learned operators may not represent a proper probability distribution, since there is a possibility that one transition will be fit by multiple preconditions; see \cite{zpk} for further discussion. For our purposes, the probabilities serve only to filter out low-probability effects on the way to guiding planning (\secref{subsec:planning}), so this technicality can be ignored.}


In terms of taxonomy, our proposed algorithm can be seen as a bottom-up inductive logic programming (ILP) algorithm, although it falls outside of the typical problem setting considered in ILP \cite{ilp1}.
In the operator learning literature \cite{arora2018review}, a close point of comparison is the algorithm of Zettlemoyer, Pasula, and Kaelbling \cite{zpk}, who learn ``noisy deictic rules'' that can be converted into probabilistic operators.
We compare our method against learning noisy deictic rules (LNDR) in experiments, finding ours to be faster and more effective for our domains.
This difference stems from our decoupling of effect clustering, precondition learning, and parameter estimation, which are all interleaved in LNDR.

\subsection{Planning with Learned Operators}
\label{subsec:planning}

We now describe how the learned probabilistic operators can be used for \tamp{}. In \secref{sec:approach1}, we described how \emph{deterministic} operators can be used for solving \tamp{} problems.
A simple and effective approach for converting probabilistic operators into deterministic ones is \emph{all-outcome determinization}~\cite{ffreplan}, whereby one deterministic operator is created for each effect set in each probabilistic operator, with the corresponding preconditions and parameters (\figref{fig:pipeline}, third panel).
Before determinizing, we filter out effects that are highly unlikely by thresholding on a hyperparameter $p_{\text{min}}$.

In the limiting case where the dataset $\D$ is empty and we were unable to learn any operators, planning reduces to simply invoking the backtracking search on every possible plan skeleton, starting from length-1 sequences, then length-2, etc. We use this strategy as a baseline (B5) in our experiments.
Since planning is still possible even without operators, it is clear that the learned operators should be understood as providing \emph{guidance} for solving  \tamp{} problems efficiently; with more data, the guidance improves.

Our aim is not to innovate on \tamp{}, but rather to demonstrate that the operators underpinning \tamp{} methods can be learned from data.
In addition to our main \tamp{} algorithm, we also found preliminary success in using \loft{} with another popular \tamp{} system~\cite{pddlstream}, but stayed with the method described in \secref{sec:approach1} due to speed advantages.

%% file: learning_pseudocode.tex
\begin{algorithm}[t]
  \SetAlgoNoEnd
  \DontPrintSemicolon
  \SetKwFunction{algo}{algo}\SetKwFunction{proc}{proc}
  \SetKwProg{myalg}{Algorithm}{}{}
  \SetKwProg{myproc}{Subroutine}{}{}
  \SetKw{Continue}{continue}
  \SetKw{Break}{break}
  \SetKw{Return}{return}
  \myalg{\textsc{Probabilistic Operator Learning}}{
    \textcolor{blue}{\tcp{\footnotesize  Transition data for a single controller}}
    \nonl \textbf{Input:} $\D_{\pi} = \{(s_i, a_i, s_{i+1}) : a_i = \pi(\cdot, \cdot) \}$\;
    \textcolor{blue}{\tcp{\footnotesize  Cluster data by lifted effect sets}}
    \nonl $\text{clusters} \gets \textsc{ClusterLiftedEffects}(\D_{\pi})$\;
    \textcolor{blue}{\tcp{\footnotesize  Compute and save precondition sets}}
    \nonl $\text{effectsToPreconditionSets} \gets \{ \}$\;
    \nonl \For{$\D_{(\pi, \text{eff})} \in $ \text{clusters}}
    {
    \textcolor{blue}{\tcp{\footnotesize  Learn one or more precondition sets}}
    \nonl \text{effectsToPreconditionSets[eff]}$\gets \textsc{LearnPreconditionSets}(\D_{(\pi, \text{eff})}, \D_{\pi})$\;
    }
    \textcolor{blue}{\tcp{\footnotesize  Instantiate operators}}
    \nonl $\text{operators} \gets \emptyset$\;
    \nonl \For{each seen precondition set $\textnormal{pre}$}
    {
    \textcolor{blue}{\tcp{\footnotesize  Effects with these preconditions}}
    $\text{effs} \gets \{ \text{eff} : \text{pre} \in \text{effectsToPreconditionSets[eff]}\}$\;
    \text{operator} $\gets$ \textsc{MakeOperator}($\pi$, pre, effs)\;
    \text{operators}.add(\text{operator})\;
    }
    \textcolor{blue}{\tcp{\footnotesize Estimate effect probabilities}}
    \nonl \For{\text{operator} $\in$ \text{operators}}
    {
    $\textsc{EstimateParameters}(\text{operator}, \D_{\pi})$\;
    }
    \nonl \Return \text{operators}\;
    }
\caption{\small{Algorithm for learning probabilistic operators for a given controller $\pi$. See \secref{subsec:learnndrs} for details.}
}
\label{alg:learning}
\end{algorithm}

%% file: experiments.tex
\section{Experiments}
\label{sec:experiments}

\subsection{Experimental Setup}

\emph{Domains.} 
We conduct experiments in three domains.

\begin{figure*}[t]
  \centering
  \noindent
    \includegraphics[width=\textwidth]{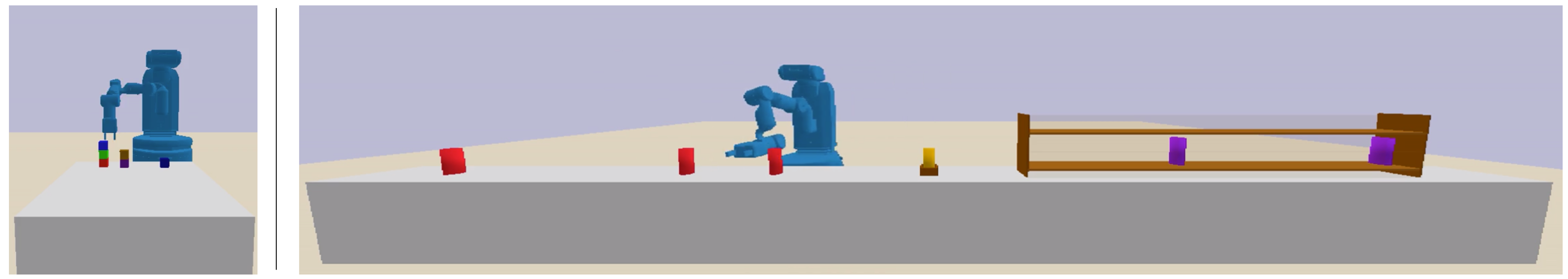}
    \caption{Snapshots from the ``Blocks'' domain (left) and the ``Painting'' domain (right). See \secref{sec:experiments} for details.}
  \label{fig:envs}
\end{figure*}

``Cover'' is a relatively simple domain in which colored blocks and targets with varying width reside along a 1-dimensional line. The agent controls a gripper that can pick and place blocks along this line, and the goal is to completely cover each target with the block of the same color. The agent can only pick and place within fixed ``allowed regions'' along the line. Because the targets and blocks have certain widths, and because of the allowed regions constraint, the agent must reason in advance about the future placement in order to decide how to grasp a block.
There are two object types (\texttt{block} and \texttt{target}), and the predicates are \texttt{Covers(?block, ?target)}, \texttt{Holding(?block)}, and \texttt{HandEmpty()}. There are two controllers, \texttt{Pick(?block, ?loc)} and \texttt{Place(?targ, ?loc)}, where the second parameter of each is a continuous location along the line.
The samplers for both actions choose a point uniformly from the allowed regions. 
We evaluate the agent on 30 randomly generated problems per seed, with average optimal plan length 3.

``Blocks'' is a continuous adaptation of the classical blocks world planning problem. A robot uses its gripper to interact with blocks on a tabletop and must assemble them into various towers. See \figref{fig:envs} for a snapshot. All geometric reasoning, such as inverse kinematics and collision checking, is implemented through PyBullet~\cite{pybullet}. The predicates are \texttt{On(?b1, ?b2)}, \texttt{OnTable(?b)}, \texttt{Clear(?b)}, \texttt{Holding(?b)}, and \texttt{HandEmpty()}. There are three controllers, \texttt{Pick(?b)}, \texttt{Stack(?b)}, and \texttt{PutOnTable(?loc)}, where \texttt{PutOnTable} is parameterized by a continuous value describing where on the table to place the currently held block. This place location must be chosen judiciously, because otherwise it could adversely affect the feasibility of the remainder of a plan skeleton. The sampler for the \texttt{PutOnTable} controller is naive, randomly sampling a reachable location on the table while ignoring obstacles.
We evaluate the agent on 10 randomly generated problems per seed, with average optimal plan length 8.

``Painting'' is a challenging, long-horizon robotic planning problem, in which a robot must place objects at target positions located in either a shelf or a box. Before being placed, objects must first be washed, dried, and painted with a certain color. Objects can be randomly initialized to start off clean or dry. To place into the shelf, the robot must first side-grasp the object due to the shelf's ceiling; similarly with top-grasping for the box.
This introduces a dependency between the grasp parameter and the feasibility of placing up to four timesteps later.
See \figref{fig:envs} for a snapshot, where the yellow object is in the box and the purple objects are in the shelf. All geometric reasoning is implemented through PyBullet~\cite{pybullet}. 
There are 14 predicates: \texttt{OnTable}, \texttt{Holding}, \texttt{HoldingSide}, \texttt{HoldingTop}, \texttt{InShelf}, \texttt{InBox}, \texttt{IsDirty}, \texttt{IsClean}, \texttt{IsDry}, \texttt{IsWet}, \texttt{IsBlank}, \texttt{IsShelfColor}, \texttt{IsBoxColor}, all parameterized by a single \texttt{?obj}, and \texttt{HandEmpty()}.   
There are five controllers, \texttt{Pick(?obj, ?base, ?grip)}, \texttt{Place(?base, ?grip)}, \texttt{Wash(?obj, ?effort)}, \texttt{Dry(?obj, ?effort)}, and \texttt{Paint(?color)}.
Here, \texttt{?effort} and \texttt{?color} are continuous values in $\mathbb{R}$, and \texttt{?base} and \texttt{?grip} are continuous values in $\mathbb{R}^3$ denoting base and end effector positions that the controller should attempt to go to before executing the pick or place.
The \texttt{Pick} sampler randomly returns a top grasp or a side grasp.
The \texttt{Place} sampler is bimodal: with probability 0.5, it samples a random placement in the shelf; otherwise, it samples a random placement in the box.
As in Blocks, the placement samplers are naive with respect to potential collisions.
The samplers for \texttt{Wash} and \texttt{Dry} are degenerate, returning exactly the appropriate \texttt{?effort} required to wash or dry the object respectively.
The \texttt{Paint} sampler randomly returns the shelf color or the box color.
We evaluate the agent on 30 randomly generated problems per seed, with average optimal plan length 31.

\begin{figure*}[t]
  \centering
  \noindent
    \includegraphics[width=\textwidth]{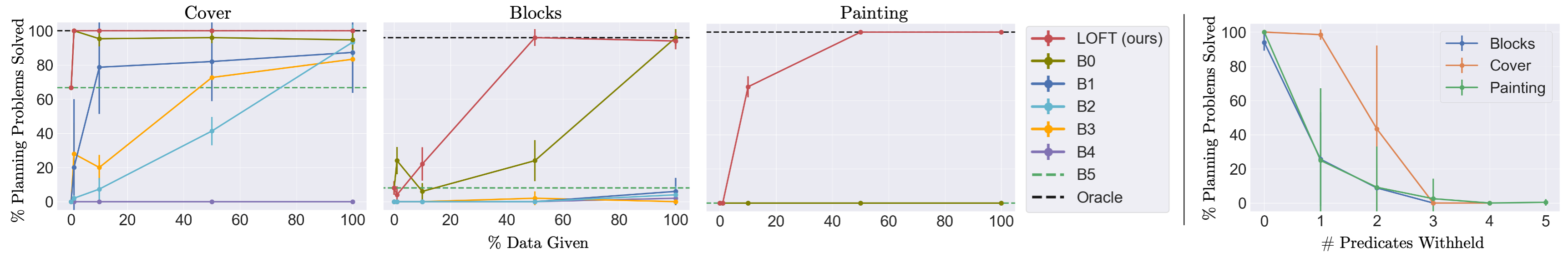}
    \caption{\emph{Left three:} For each of our domains, the percent of planning problems solved within a timeout, as a function of the amount of data provided to the approach. Each point is a mean over 5 seeds. Dotted lines indicate non-learning approaches (B5 and Oracle). Planning timeout was set to 1 second for Cover and 10 seconds for Blocks and Painting. In all domains, \loft{} performs extremely well, reaching better planning performance than other approaches with less data. \emph{Right:} Predicate ablation experiment, showing problems solved by \loft{} as a function of the number of predicates withheld. Each point is a mean over 25 seeds; for each seed, we randomly select predicates to remove. \loft{} is somewhat robust to a few missing predicates, but generally relies on being given a complete set to perform very well.}
  \label{fig:results}
\end{figure*}

\emph{Methods Evaluated.} We evaluate the following methods.
\begin{itemize}
    \item \loft{}: Our full approach.
    \item Baseline B0: We use \loft{} but replace our \algref{alg:learning} for probabilistic operator learning with LNDR~\cite{zpk}, a popular greedy algorithm for learning noisy rules.
    \item Baseline B1: Inspired by Kim and Shimanuki~\cite{kim2020learning}, we train a graph neural network (GNN) that represents a Q-function for high-level search. The model takes as input a symbolic state, an action template, and the goal, and outputs a Q-value for that action template. To use the model, we perform tree search as described in \cite{kim2020learning}: the Q-function is used to select the best action template, and we sample $W$ (a width parameter) continuous values for calling the simulator to produce successor states. A major difference from their work is that we are not doing online learning, so we perform fitted Q-iteration with the given dataset $\D$ in order to train the GNN.
    \item Baseline B2: Same as B1, but the low-level state $x$ is also included in the input to the GNN.
    \item Baseline B3: Inspired by Driess et al.~\cite{driess2020deep}, we train a recurrent GNN that predicts an entire plan skeleton conditioned only on the initial low-level state, initial symbolic state, and goal. This recurrent model sequentially predicts the next action template to append onto the skeleton. For each skeleton, we use our backtracking search method (\secref{sec:approach1}) to attempt to optimize it.
    \item Baseline B4: We train a raw GNN policy that maps low-level states $x$ to actions $a$. The GNN outputs a discrete choice of which controller to use, and values for both the discrete and continuous arguments of that controller. This baseline does not make use of the samplers.
    \item Baseline B5: As discussed in \secref{subsec:planning}, we run our planning algorithm with no operators, which reduces to trying to optimize every possible plan skeleton. This is the limiting case of \loft{} where the dataset $\D=\emptyset$.
    \item Oracle: We use our planner with good, hand-written operators for each domain. This method represents an upper bound on the performance we can get from \loft{}.
\end{itemize}

\emph{Data Collection.}
In all domains, we use the same data collection strategy. First, we generate a set of 20 problems from the domain that are smaller (with respect to the number of objects) than the ones used for evaluation. Then, we use an oracle planner to produce demonstrations of good behavior in these smaller problems. Finally, we collect ``negative'' data (which is important for learning preconditions) by, $K$ times, sampling a random state $x$ seen in the demonstrations, taking a random action $a$ from that state, and seeing the resulting $x' = f(x, a)$, where $f$ is our low-level simulator. We use $K=100$ for Cover and Blocks, and $K=2500$ for Painting.

The number of transitions $|\D|$ is 126 for Cover, 152 for Blocks, and 2819 for Painting. All learning-based methods (\loft{} and Baselines B0-B4) receive the exact same dataset.

\emph{Experimental Details.}
We use $p_{\text{min}}=0.001$ and $N_\text{samples}=10$ for all domains. We use a planning timeout of 1 second for Cover and 10 seconds for Blocks and Painting.
For all GNN models, we preprocess literals to be arity 1 or 2 by either adding a dummy argument (if arity 0) or splitting into multiple literals (if arity $>2$).
All GNNs are standard encode-process-decode architectures \cite{gnn}, where node and edge modules are fully connected neural networks with one hidden layer of dimension 16, ReLU activations, and layer normalization.
Message passing is performed for $K=3$ iterations.
Training uses the Adam optimizer with learning rate 0.001 and batch size 16.
For B1 and B2, we use search width $W=1$ for Cover and Blocks, and $W=3$ for Painting; we use 5 iterations of fitted Q-iteration for Cover, and 15 for Blocks and Painting; and we train for 250 epochs per iteration. For B3 and B4, we train for 1000 epochs.

\subsection{Results and Discussion}

\input{tables}

See \figref{fig:results} (left three plots) for our main set of results, which show planning performance of each method as a function of the amount of data given.
To assess performance with 50\% of data given, for example, we train from scratch using only 50\% of transitions randomly sampled from the full dataset $\D$.
In all three domains, \loft{} achieves the performance of the oracle with enough data.
In Painting, none of the baselines are able to solve even a single planning problem within 10 seconds. Furthermore, across all domains, \loft{} is more data-efficient than all learning-based baselines (especially the GNN ones), since it does not require as much data on average to learn probabilistic operators as it does to train a neural network.

The difference between \loft{} and B0 (\loft{} but with LNDR) is noteworthy: B0 often performs much worse.
Inspecting the operators learned by LNDR, we find that the learning method is more liable to get stuck in local minima.
For example, in the Painting domain, LNDR consistently uses one operator with the \texttt{Holding} predicate instead of two operators, one with \texttt{HoldingSide} and another with \texttt{HoldingTop}.
Our learning method is able to avoid these pitfalls primarily because the lifted effect clustering is decoupled from precondition learning and parameter estimation.
Our method learns effect sets once and does not revisit them until the operators are determinized during planning; in contrast, LNDR constantly re-evaluates whether to keep or discard an effect set in the course of learning operators.

Comparing B1 and B2 shows that model-free approaches can actually suffer from inclusion of the low-level state as input to the network, since this inclusion may make the learning problem more challenging. \loft{}, on the other hand, does not use the low-level states to learn operators; this may be a limitation in situations where the predicates are not adequate for learning useful operators. Nevertheless, one could imagine combining model-free approaches like B1, B2, and B3 with \loft{}: a learned Q-function could serve as a heuristic in our high-level $\text{A}^*$ search.

B4, the raw GNN policy, performs especially poorly in all domains since it does not make use of the samplers. This suggests that direct policy learning without hand-written samplers is challenging in many \tamp{} domains of interest.

\loft{} outperforms B5 (planning without operators) given enough data.
However, in Blocks, we see that with only a little data, \loft{} actually performs slightly \emph{worse} than B5, likely because with little data, the operators that are learned are quite poor and provide misleading guidance.

\tabref{tab:trainingtimes} reports training times for all methods. 
We see that \loft{} trains extremely quickly --- many orders of magnitude faster than the GNN baselines, which generally do not perform nearly as well as \loft{} in our domains.
This further confirms that learning symbolic operators is a useful and practical way of generating guidance for \tamp{} planners.

Finally, in two additional experiments, we measure the impact of ablating predicates on the performance of \loft{} (\figref{fig:results}, rightmost plot) and test the importance of using classical heuristics in the high-level A* search (\tabref{tab:heuristicablation}).
We can conclude from \figref{fig:results} that \loft{} has some robustness to missing predicates: even with up to 3 predicates withheld, \loft{} remains the only non-oracle approach that solves any planning problems in Painting. However, performance deteriorates quickly; this is because we are using the operators in A* search with the hAdd heuristic, which is highly sensitive to missing preconditions or effects. \tabref{tab:heuristicablation} is an ablation study that shows the importance of using hAdd: our results are significantly worse if we instead use a blind heuristic, that is, a heuristic which is 0 everywhere. This speaks to the benefits of the fact that \loft{} learns PDDL-style operators.

%% file: tables.tex
\begin{table*}
\centering
\begin{tabular}{ |p{2cm}||c|c|c|c|c|c|  }
 \hline
 \multicolumn{7}{|c|}{Training Times (seconds)} \\
 \hline
\emph{Domain} & \textbf{\loft{} (ours)} & \textbf{B0} & \textbf{B1} & \textbf{B2} & \textbf{B3} & \textbf{B4}\\
 \hline
 Cover   & 0.13 (0.07)  & 0.47 (0.03)  & 131 (17)  & 145 (21) & 497 (520) & 127 (174)  \\
 \hline
 Blocks   & 0.12 (0.07)  & 6.6 (0.42)  & 490 (48)  & 554 (72)  & 371 (122)  & 106 (146)  \\
 \hline
 Painting   & 16 (0.91)  & 328 (21)  & 23704 (1033)  & 24326 (490)  & 5371 (3904)  & 816 (361)  \\
 \hline
\end{tabular}
\caption{Each entry shows the mean (standard deviation) training time in seconds over 5 seeds. Both \loft{} and B0 train orders of magnitude faster than B1-B4. \loft{} is also one order of magnitude faster than B0. B5 is excluded because it is not learning-based.}
\label{tab:trainingtimes}
\end{table*}

\begin{table}
\centering
\footnotesize
\begin{tabular}{ |p{2cm}||c|c|  }
 \hline
 \multicolumn{3}{|c|}{Heuristic Ablation Results} \\
 \hline
\emph{Domain} & \textbf{\loft{} w/ hAdd} & \textbf{\loft{} w/ blind} \\
 \hline
 Cover   & 100 (0)  & 100 (0)  \\
 \hline
 Blocks   & 94 (5)  & 42 (12)  \\
 \hline
 Painting   & 100 (0)  & 0 (0)  \\
 \hline
\end{tabular}
\caption{An ablation of the hAdd heuristic used in our method, where we replace it with a blind heuristic (always 0) that does not leverage the PDDL structure of the operators. Each entry shows the mean (standard deviation) percent of planning problems solved over 5 seeds. These results suggest that classical planning heuristics like hAdd are especially critical in our harder domains.}
\label{tab:heuristicablation}
\end{table}

%% file: conclusion.tex
\section{Conclusion}
\label{sec:conclusion}

We addressed the problem of learning operators for \tamp{} in hybrid robotic planning problems. The operators guide high-level symbolic search, making planning efficient. Experiments on long-horizon planning problems demonstrated the strength of our method compared to several baselines, including graph neural network-based model-free approaches.

A key future research direction is to relax the assumption that predicates are given. While assuming given predicates is standard in learning for \tamp{}, this is a severe limitation, since the quality of these predicates determines the quality of the abstraction. It would be useful to study what characterizes a good predicate, toward designing a predicate learning algorithm. Another useful future direction would be to learn the hybrid controllers that we are currently given, perhaps by pretraining with a reinforcement learning algorithm.

%% file: acknowledgements.tex
\section*{Acknowledgements}
We gratefully acknowledge support from NSF grant 1723381; from AFOSR grant FA9550-17-1-0165; from ONR
grant N00014-18-1-2847; from the Honda Research Institute; from MIT-IBM Watson Lab; and from
SUTD Temasek Laboratories. Rohan and Tom are supported by NSF Graduate Research Fellowships.
Any opinions, findings, and conclusions or recommendations expressed in this material are those of
the authors and do not necessarily reflect the views of our sponsors.